\documentclass[10pt,twocolumn]{article}

\usepackage[margin=1in]{geometry}

\usepackage{amsmath,amssymb,amsfonts,amsthm}

\usepackage{graphicx}

\usepackage{array}

\usepackage[numbers]{natbib}  
\usepackage{caption} 
\usepackage{times}  
\usepackage{helvet}  
\usepackage{courier}  
\usepackage[hyphens]{url}  

\usepackage{amsfonts}
\usepackage{makecell}
\usepackage{mathtools}
\usepackage{amsthm}
\usepackage{thmtools}
\usepackage{algorithm}
\usepackage[noend]{algpseudocode}
\usepackage{xspace}
\usepackage{siunitx}
\usepackage{enumitem} 
\usepackage{tabularx} 
\usepackage{tcolorbox}
\usepackage{booktabs}

\usepackage{multirow}
\usepackage{diagbox}

\newtheorem{definition}{Definition}

\usepackage{cleveref}

\crefname{figure}{Fig.}{Figs.}
\Crefname{figure}{Fig.}{Figs.}

\crefname{table}{Table}{Tables}
\Crefname{table}{Table}{Tables}

\crefname{section}{Sect.}{Sects.}
\Crefname{section}{Sect.}{Sects.}

\crefname{appendix}{Appendix}{Appendices}
\Crefname{appendix}{Appendix}{Appendices}

\crefname{definition}{Def.}{Defs.}
\Crefname{definition}{Def.}{Defs.}

\crefname{algorithm}{Alg.}{Algs.}
\Crefname{algorithm}{Alg.}{Algs.}

\crefname{equation}{Formula}{Formulas}
\Crefname{equation}{Formula}{Formulas}
\creflabelformat{equation}{#2#1#3}

\newcommand{\woodelf}{\textsc{Woodelf}\xspace} 
\newcommand{\pltreeshap}{\textsc{PLTreeSHAP}\xspace} 
\newcommand{\hdwoodelf}{\textsc{WoodelfHD}\xspace} 
\newcommand{\shap}{\texttt{shap}\xspace}
\newcommand{\fasttreeshap}{\textsc{FastTreeSHAP}\xspace}
\newcommand{\lineartreeshap}{\textsc{Linear TreeSHAP}\xspace}
\newcommand{\treegradshap}{\textsc{TreeGrad-SHAP}\xspace}
\newcommand{\lightgbm}{\textsc{LightGBM}\xspace}
\newcommand{\gputreeshap}{\textsc{GPUTreeSHAP}\xspace}

\algnewcommand\algorithmicyield{\textbf{yield}}
\algnewcommand\Yield[1]{\algorithmicyield\ #1}

\title{WOODELF-HD: Efficient Background SHAP for High-Depth Decision Trees}

\author{
Ron Wettenstein$^{1}$, Alexander Nadel$^{2}$, Udi Boker$^{1}$\\[0.5em]
$^{1}$Reichman University, Herzliya, Israel\\
$^{2}$Faculty of Data and Decision Sciences, Technion, Haifa, Israel\\
ron.wettenstein@post.runi.ac.il, 
alexandernad@technion.ac.il
}
\date{}

\begin{document}

\maketitle

\begin{abstract}

Decision-tree ensembles are a cornerstone of predictive modeling, and \emph{SHAP} is a standard framework for interpreting their predictions. Among its variants, \emph{Background SHAP} offers high accuracy by modeling missing features using a background dataset. Historically, this approach did not scale well, as the time complexity for explaining $n$ instances using $m$ background samples included an $O(mn)$ component. Recent methods such as \woodelf and \pltreeshap reduce this to $O(m+n)$, but introduce a preprocessing bottleneck that grows as $3^D$ with tree depth $D$, making them impractical for deep trees.

We address this limitation with \hdwoodelf, a \woodelf extension that reduces the $3^D$ factor to $2^D$. The key idea is a Strassen-like multiplication scheme that exploits the structure of \woodelf matrices, reducing matrix–vector multiplication from $O(k^2)$ to $O(k \log k)$ via a fully vectorized, non-recursive implementation. In addition, we merge path nodes with identical features, reducing cache size and memory usage.

When running on standard environments, \hdwoodelf enables exact Background SHAP computation for trees with depths up to $21$, where previous methods fail due to excessive memory usage. For ensembles of depths $12$ and $15$, it achieves speedups of $33\times$ and $162\times$, respectively, over the state-of-the-art.

\end{abstract}

\subsection*{Links}
\noindent
\textbf{The \woodelf and \hdwoodelf Python Package} -- \\
https://github.com/ron-wettenstein/woodelf

\vspace{0.2em}

\noindent
\textbf{Experimental Results Interactive Report} --- \\
https://ron-wettenstein.github.io/TreeBranchMarks/ \\
benchmarks/reports/WoodelfhdMainExperiment.html

\vspace{0.2em}

\noindent
\textbf{Experiment Notebooks} --- \\
https://github.com/ron-wettenstein/WoodelfExperiments

\begin{table*}[t]
\small
\centering
\def\arraystretch{1.6}
\begin{tabularx}{\textwidth}{X|c|c|c|c}
\textbf{Task} &
\textbf{\shap Package} &
\textbf{PLTreeSHAP} &
\textbf{Original \woodelf} &
\textbf{\hdwoodelf} \\
\hline

BG SHAP &
$O(nmTLD)$ &
$O(mTL \! + \! nT3^DD)$ &
$O(mTL \! + \! nTLD \! + \! TL3^DD)$ &
$O(mTL \! + \! nTLD \! + \! TL2^DD^2)$ \\

BG SHAP IV &
$O(nmTLDf)$ &
$O(mTL \! + \! nT3^DD^2)$ &
$O(mTL \! + \! nTLD^2 \! + \! TL3^DD^2)$ &
$O(mTL \! + \! nTLD^2 \! + \! TL2^DD^3)$ 

\end{tabularx}
\caption{\label{complexity_table}
Time complexities for Background SHAP (BG SHAP) and Background SHAP interaction values (BG SHAP IV). The table presents the complexities of the \shap Python package, \pltreeshap, original \woodelf (as presented in~\cite{woodelf}) and \hdwoodelf (as presented in this paper). For notation, see~\cref{fig:notations}. 
The \shap complexity depends on $O(nm)$, which becomes impractical for large background datasets. Among the methods with $O(n+m)$ dependence, \hdwoodelf achieves the best asymptotic complexity, depending on $2^D$ rather than $3^D$.
}
\end{table*}

\section{Introduction}

Decision tree ensembles such as XGBoost~\cite{xgboost_paper}, Random Forest~\cite{random_forrest}, 
and CatBoost~\cite{catboost} are widely used predictive models for classification and regression. With their increasing adoption, substantial attention has been devoted to explaining their predictions. One common perspective frames explanation as the fair distribution of predictive credit among input features. This viewpoint is formalized by SHAP~\cite{unified_approach_to_interpreting_models}, which models prediction explanation as a cooperative game and assigns feature attributions using Shapley values~\cite{original_shapley_paper}. Beyond individual attributions, SHAP also extends to Shapley interaction values, which capture pairwise feature interactions~\cite{PB_shap_and_banzhaf}.

Computing Shapley values is \#P-hard for certain games~\cite{shapley_calculation_is_npc} and for common model classes such as neural networks and logistic regression~\cite{tractability_shap_explanations, complexity_of_SHAP}. Fortunately, exact and efficient computation is possible for decision tree ensembles~\cite{explainable_ai_trees} as well as in other settings, including certain classes of Boolean circuits~\cite{arenas2021tractabilityshapscorebasedexplanationsdeterministic} and tuples in query answering~\cite{shapley_on_queries}.

SHAP has two commonly used variants. \emph{Background SHAP} uses a background dataset to accurately capture the model prediction when some features are missing, while \emph{Path-Dependent SHAP} estimates this behavior based on node-coverage statistics. While our proposed solution applies to both variants, we focus primarily on Background SHAP, the more accurate approach, where we improve the state-of-the-art complexity.

The most widely used implementation of SHAP is the \shap Python package, which supports both CPU~\cite{explainable_ai_trees} and GPU~\cite{GPUTreeShap} computation. \shap's key drawback is its quadratic complexity for Background SHAP: The complexity of explaining the predictions of $n$ instances with a background dataset of size $m$ includes an $O(mn)$ component. The \pltreeshap algorithm~\cite{linear_background_shap} reduces this to $O(m+n)$ by preprocessing the decision tree and the background dataset. 

For Path-Dependent SHAP, \fasttreeshap~\cite{fast_tree_shap} was the first algorithm to outperform the original \shap implementation. More recent advances include \lineartreeshap~\cite{lineartreeshap}, which improves the theoretical complexity but suffers from numerical instability (see \cite{treegradranker}, Section 5), and \treegradshap~\cite{treegradranker}, which builds on this line of work and significantly reduces these stability issues.

Finally, recent work introduced \woodelf, a unified and efficient algorithm for computing Shapley values on decision tree ensembles~\cite{woodelf}. \woodelf computes both Path-Dependent and Background SHAP in a numerically stable manner and, like \pltreeshap, computes Background SHAP with complexity that scales as $O(m+n)$. It is significantly faster than \shap, \pltreeshap, and \fasttreeshap, especially on large datasets. \woodelf is GPU-friendly and available as a Python package. In this paper, we extend \woodelf to efficiently support high-depth trees.

\subsection{Our Contributions}

\begin{table}[t]
\small
\centering
\def\arraystretch{1.5}
\begin{tabular}{l|c|c}
\textbf{Task} &
\textbf{Original \woodelf} &
\textbf{\hdwoodelf} \\
\hline
\rule{0pt}{4ex}

PD SHAP &
\makecell{$O(\boldsymbol{L3^D D}\,+$ \\ $n(f+L))$} &
\makecell{$O(\boldsymbol{2^D D}\,+$ \\ $n(f+D))$} \\\rule{0pt}{6ex}

BG SHAP &
\makecell{$O(\boldsymbol{L3^D D}\,+$ \\ $n(f+L) +\, mL)$} &
\makecell{$O(\boldsymbol{2^D D}\,+$ \\ $n(f+D) +\, mD)$} \\\rule{0pt}{6ex}

PD SHAP IV &
\makecell{$O(\boldsymbol{L3^D D^2}\,+$ \\ $n(f^2+L))$} &
\makecell{$O(\boldsymbol{2^D D^2}\,+$ \\ $n(f^2+D))$} \\\rule{0pt}{6ex}

BG SHAP IV &
\makecell{$O(\boldsymbol{L3^D D^2}\,+$ \\ $n(f^2+L) +\, mL)$} &
\makecell{$O(\boldsymbol{2^D D^2}\,+$ \\ $n(f^2+D) +\, mD)$} \\

\end{tabular}

\caption{\label{space_complexity_table}
Space complexities of the original \woodelf~\cite{woodelf} and \hdwoodelf, considering both the Path-Dependent (PD) and Background (BG) variants for Shapley values (SHAP) and Shapley interaction values (SHAP IV). For notation, see~\cref{fig:notations}.
}
\end{table}

A key limitation of both \woodelf and \pltreeshap is their poor scalability to deep trees. Their time and space complexities have a $(3^D D)$ component, where $D$ is the tree depth. In practice, this makes the algorithms difficult to use once the maximum depth exceeds about 12, as runtimes increase sharply and memory requirements typically become prohibitive.

Although limiting tree depth is a common way to control overfitting~\cite{Breiman1984}, deep trees are still frequently used in practice. For example, \lightgbm~\cite{LightGBM} employs a leaf-wise (depth-first) growth strategy by default and does not impose an explicit depth limit. This strategy expands splits along a single path before broadening the tree, often producing trees with a large maximum depth.

\begin{figure}[t]
    \centering
    \begin{tcolorbox}
        \label{box:notation}
        $\boldsymbol{f}$: no.\ of features;
        $\boldsymbol{n}$: no.\ of consumer data rows;
        $\boldsymbol{m}$: no.\ of background data rows;
        $\boldsymbol{T}$: no.\ of trees;
        $\boldsymbol{D}$: max depth;
        $\boldsymbol{L}$: max leaves per tree.
    \end{tcolorbox}
    \caption{\label{fig:notations} Notation.}
\end{figure}
We introduce \hdwoodelf (\woodelf for High Depth), an improved algorithm that enhances the state-of-the-art complexity for Background SHAP by replacing the $3^D D$ component with $2^D D^2$ in both time and space. \cref{complexity_table,space_complexity_table} summarize the resulting complexity reductions. Notation is given in \cref{fig:notations}.

These improvements also extend to Path-Dependent SHAP. While the theoretical state-of-the-art remains $O(nTLD)$ as established by \lineartreeshap, our approach often outperforms both \treegradshap and \shap in practical applications involving large-scale datasets. These gains are driven by \hdwoodelf vectorized implementations and efficient preprocessing.

\hdwoodelf significantly outperforms the state-of-the-art in computing Background SHAP on deep trees and extends the range of depths that can be handled in practice. On a 50GB RAM machine, \hdwoodelf runs efficiently up to depth 21, whereas \woodelf fails due to excessive memory usage beyond depth 15. At depth 12, on the IEEE-CIS dataset, \hdwoodelf computes the required Shapley values on a large dataset in under 2 minutes, while \woodelf requires 58 minutes for the same task (see \cref{tab:depth_running_time} in \cref{sec:experimental_results}).

These advancements extend the applicability of the \woodelf approach to many practical settings. For example, a large-scale empirical study across 38 datasets found that the optimal maximum depth for XGBoost is around 13~\cite{Probst2019Tunability}. Such depths already exceed the practical limit of \woodelf and \pltreeshap, but are efficiently supported by \hdwoodelf.

\subsection{Key Ideas and Optimizations}

The complexity improvements summarized in \cref{complexity_table,space_complexity_table} are achieved through several key ideas:
\begin{enumerate}
    \item We run the entire \woodelf pipeline on each leaf independently, rather than on each tree. This change significantly reduces space complexity.
    \item We merge nodes that split on the same feature, ensuring that features are unique along each root-to-leaf path. This enables a compact caching mechanism that accelerates preprocessing.
    \item A major bottleneck in the \woodelf preprocessing phase is a vector--matrix multiplication between a sparse matrix $M$ and a vector $f$. The matrix $M$ has $2^D$ rows, $2^D$ columns, and only $3^D$ non-zero entries. The vector $f$ has length $2^D$. In the original work~\cite{woodelf}, this multiplication is performed using sparse methods in $O(3^D)$ time. Here, we exploit the special structure of $M$, which arises from the linearity of Shapley values, and propose a Strassen-like vector--matrix multiplication scheme with time complexity $O(2^D D)$.
    \item We further observe that this new multiplication scheme depends only on the secondary diagonal of $M$. We therefore compute and store only this diagonal. Combined with the Strassen-like scheme, this reduces the multiplication time and space complexity from $O(3^D)$ to $O(2^D D)$.
    \item A naive implementation of the Strassen-like scheme requires $O(2^D)$ recursive calls. Instead, we derive a fully vectorized version that uses only $O(D)$ operations on $2^D$-sized vectors, eliminating recursion and greatly improving performance.
\end{enumerate}

Overall, our approach is vectorized using Numpy~\cite{numpy_article}, utilizes SIMD (Single-Instruction Multiple-Data~\cite{SIMD}), is GPU-friendly, and is implemented in pure Python. The method is integrated into the \texttt{Woodelf} Python package, where it serves as the default implementation.

\section{The \woodelf Algorithm}
\label{sec:woodelf_overview}

This section briefly reviews the \woodelf algorithm~\cite{woodelf}, which our method builds upon. We begin with a short description of decision trees and root-to-leaf paths.

A decision tree is a rooted binary tree in which each node is either an internal node or a leaf. Each leaf is associated with a weight. An internal node $n$, with left child $n.l$ and right child $n.r$, specifies a split $n.\textit{split}$ of the form $n.\textit{feature} < n.\textit{threshold}$: if the split holds, traversal proceeds to $n.l$, and otherwise to $n.r$. A \emph{root-to-leaf path} for a leaf $l$ is the unique path from the root to $l$.

The first observation underlying \woodelf is that SHAP computation at a given leaf depends only on the splits along its root-to-leaf path. For each consumer row and background row, only the split outcomes along this path influence the resulting Shapley values. \woodelf encodes these outcomes as a \emph{decision pattern} (see \cref{def:decision_pattern}).

\begin{definition}[Decision Pattern {\cite[Definition 7]{woodelf}}]
\label{def:decision_pattern}
Given a decision tree $T$, a leaf $l$, their root-to-leaf path $(n_1 \equiv r_T, n_2, \dots, n_{D-1}, l)$, and feature values $c = (c_1, \dots, c_h) \in \mathbb{R}^h$, the \emph{decision pattern} $p$ is a binary sequence of length $D-1$. Its $i$-th bit is defined as:
\[
p[i] =
\begin{cases}
1 & \text{if } (n_i.\textit{split}(c) = \text{True}) \land (n_i.\textit{left} = n_{i+1}), \\[6pt]
1 & \text{if } (n_i.\textit{split}(c) = \text{False}) \land (n_i.\textit{right} = n_{i+1}), \\[6pt]
0 & \text{otherwise}.
\end{cases}
\]
\end{definition}

To efficiently compute decision patterns, the \emph{CalcDecisionPatterns} algorithm (\cref{alg:calc_decision_patterns} in~\cref{appendix:DecisionPatternsGenerator}) is used. 

We now turn to cooperative games and their representation as pseudo-Boolean formulas. A cooperative game is defined by a characteristic function $C : 2^{[N]} \to \mathbb{R}$, which maps each subset of players to the profit obtained when only those players participate. Equivalently, this function can be represented as a pseudo-Boolean function $C : \{0,1\}^{|N|} \to \mathbb{R}$, where each player corresponds to a Boolean variable. The profit of a subset is then given by evaluating this function with participating players set to $\mathrm{True}$ and all others set to $\mathrm{False}$.

\begin{figure}[t]
    \centering
    \includegraphics[width=0.7\linewidth]{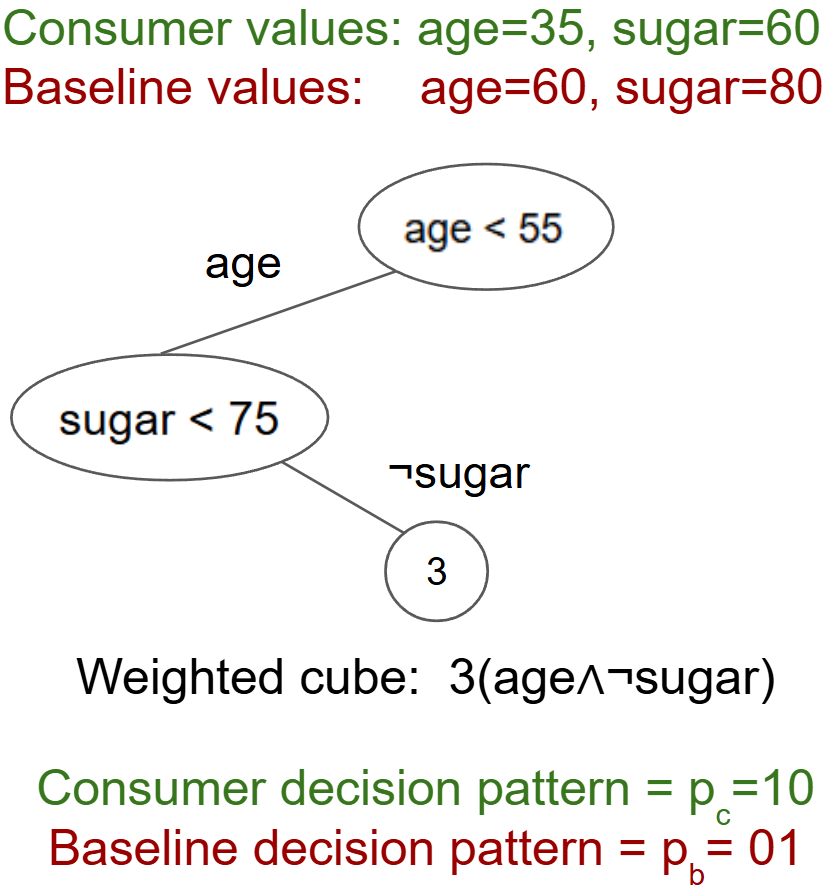}
    \caption{A consumer row, baseline row and root-to-leaf path. The figure displays the consumer decision pattern (\cref{def:decision_pattern}), the baseline decision pattern, and the weighted cube that matches these patterns. The cube is constructed according to the procedure in lines~\ref{line:m2_creation}-\ref{line:m4_creation} of \cref{alg:fill_wdnf_table}.}
    \label{fig:cube_and_decision_patterns_example}
\end{figure}

A \emph{literal} is a Boolean variable $x_i$ or its negation $\neg x_i$, and a \emph{cube} is a conjunction of literals. For example, $(x_2 \land x_3\land \neg x_5)$ is a cube. The \woodelf algorithm observes that the characteristic function induced by a single leaf, consumer, and background row can be represented as a weighted cube (see~\cref{fig:cube_and_decision_patterns_example} for an example). 

Given the features along a root-to-leaf path, the procedure \emph{MapPatternsToCube} (\cref{alg:fill_wdnf_table}) constructs a matrix that maps each pair of consumer and background decision patterns (row and column indices, respectively) to the cube they induce at that leaf. For a path with $D$ unique features, this matrix has $2^D$ rows, $2^D$ columns, and $3^D$ non-empty entries. See an example of such a matrix in \cref{fig:m_cube_on_path_example}.

\cite{woodelf} further provides a simple formula for computing Shapley values for games represented by weighted cubes. For each feature along the root-to-leaf path, \woodelf applies its Shapley-values formula to all cubes in the matrix constructed by \emph{MapPatternsToCube}. The resulting matrix for feature $i$ maps each pair of consumer and background decision patterns to their corresponding Shapley value for feature $i$. We refer to these matrices as the \emph{$M$ matrices}. They play a central role in this work: one of our key contributions is to analyze their structure and leverage it to accelerate matrix–vector multiplication.

\begin{table}[t]
\centering
\renewcommand{\arraystretch}{1.4}
\begin{tabular}{c|c|c|c|c|c}

\diagbox[width=3.5em]{$p_c$}{$p_b$} & 00 & 01 & 10 & 11 \\\hline
00 & $\emptyset$ & $\emptyset$ & $\emptyset$ & \makecell{$(\neg age \: \wedge $ \\ $  \neg sugar)$} \\\hline
01 &  $\emptyset$ & $\emptyset$ & \makecell{$(\neg age \: \wedge $ \\ $ sugar)$} & $(\neg age)$ \\\hline
10 &  $\emptyset$ & \makecell{$(age  \: \wedge $ \\ $ \neg sugar)$} & $\emptyset$ & $(\neg sugar)$ \\\hline
11 & \makecell{$(age \: \wedge $ \\ $ sugar)$} & $(age)$ & $(sugar)$ & $()$ \\\hline
\end{tabular}
\caption{The matrix produced by \emph{MapPatternsToCube} for the root-to-leaf path shown in 
\cref{fig:cube_and_decision_patterns_example}. The entry at row index $10$  and column index $01$ is the cube that matches the 
consumer and baseline rows in \cref{fig:cube_and_decision_patterns_example}. Applying any function that satisfies linearity, such as the Shapley value of $sugar$ or $age$, to the cubes in this table yields an $M$ matrix. These kinds of matrices are discussed in \cref{sec:strassen_mult}. }
\label{fig:m_cube_on_path_example}
\end{table}

\begin{algorithm}[t]
\caption{Map Patterns to Cube {~\cite[Alg. 2]{woodelf}}\\ {\bf Algorithm \ref{alg:fill_wdnf_table}'} Cubes In Diagonal (without line~\ref{line:m4_creation})}\
\label{alg:fill_wdnf_table}
\begin{algorithmic}[1]
\Function{MapPatternsToCube}{$\textit{features}$}
    \State $d \gets\{ 0 \mapsto \{ 0 \mapsto (\emptyset, \emptyset) \} \}$
    \For{$f  \in \textit{features}$}
        \State $d_{old} \gets d$
        \State $d \gets \{\}$
        \For{$p_c$ in $d_{old}$}
            \For{$p_b$ in $d_{old}[p_c]$}
                \State $(S^+, S^-) \gets d_{old}[p_c][p_b]$ \label{line:extract_splus_sminus}
                \State $d[2p_c \!+ \!1][2p_b \!+ \!0] \! \gets \! (S^+ \cup \{f\}, S^-)$ \label{line:m2_creation}
                \State $d[2p_c \! + \!0][2p_b \!+ \!1] \! \gets \! (S^+, S^- \cup \{f\})$ \label{line:m3_creation}
                \State $d[2p_c \! +\! 1][2p_b\! +\! 1] \! \gets \! (S^+, S^-)$ \label{line:m4_creation}
            \EndFor
        \EndFor
    \EndFor
    \State \Return $d$
\EndFunction
\end{algorithmic}
\end{algorithm}

At a high level, the \woodelf algorithm operates on a single decision tree and consists of four main steps:
\begin{enumerate}
    \item \textbf{Computing the $f$ vectors}: The algorithm finds the background decision patterns for all leaves, computes their \textit{value\_counts}, and stores the results as the $f$ vectors. \label{item:f_computation}
    \item \textbf{Computing the $M$ matrices}: For each leaf, \woodelf precomputes the Shapley values for every possible pair of consumer and background decision patterns. This is done by generating matrices using \emph{MapPatternsToCube} and applying the Shapley value function to their cubes. \label{item:M_generation}
    \item \textbf{Computing the $s$ vectors}: This is done using sparse matrix-vector multiplication, $s = M \cdot f$. For each leaf, the resulting $s$ vector maps each possible consumer pattern with its Shapley values.
    \item \textbf{Fetching the final Shapley values using the $s$ vectors}: For each leaf and consumer, compute the consumer’s decision pattern and use it with the leaf’s $s$ vector to obtain its contribution to the Shapley values. Summing over all leaves yields the consumer’s final Shapley values.
\end{enumerate}

\begin{figure*}[t]
\centering
\includegraphics[width=0.8\textwidth]{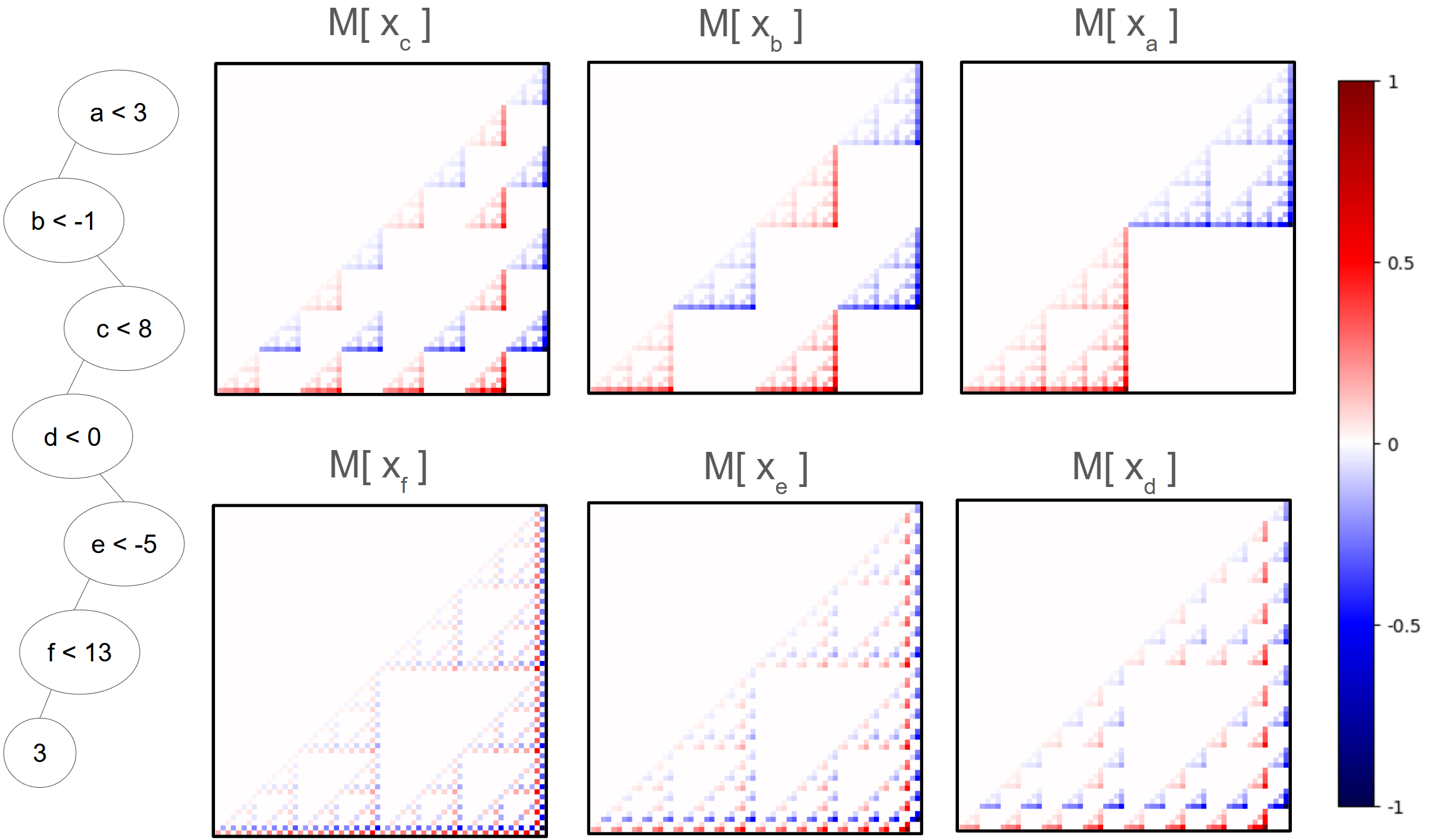} 
\caption{
The $M$ matrices corresponding to a root-to-leaf path with 6 unique features. 
The matrix $M[x_a]$ matches feature $a$ from the first split, $M[x_b]$ matches feature $b$, and so on. 
Each cell is colored according to its value: $1$ appears as dark red, $-1$ as dark blue, $0$ as white, and intermediate values are shown with opacity proportional to their magnitude.
}
\label{fig:M_plot}
\end{figure*}

\begin{figure*}[t]
\centering
\includegraphics[width=0.8\textwidth]{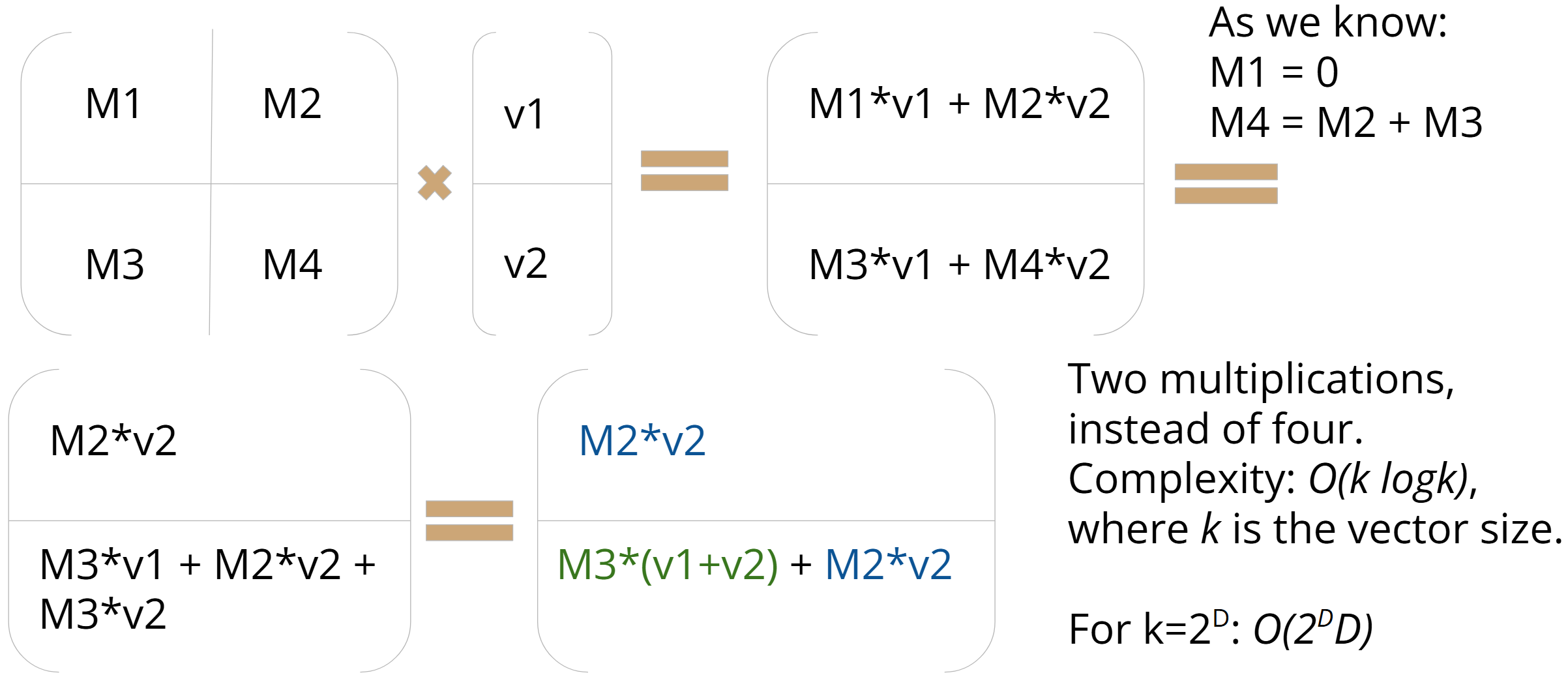} 
\caption{
Using the matrix properties to perform matrix–vector multiplication with only 2 inner multiplications instead of 4. 
This Strassen-like formulation reduces the complexity from $O(k^2)$ to $O(k \cdot \log k)$, i.e., from $O(4^{D})$ to $O(2^D D)$.
}
\label{fig:strassen_formulation}
\end{figure*}

\section{Strassen-Like Multiplication}
\label{sec:strassen_mult}

This section presents an efficient approach for computing the product $M \cdot f$ described in \cref{sec:woodelf_overview}. The method leverages structural properties of the $M$ matrices and draws inspiration from Strassen’s algorithm for fast matrix multiplication~\cite{Strassen1969}. 

For a formal definition of the $M$ matrices, see~\cref{sec:woodelf_overview}, which describes how \woodelf constructs them. Alternatively, \cref{sec:M_matix_def_over_PB_functions} defines them using only pseudo-Boolean terminology. The perspective presented in the appendix can help generalize the approach discussed in this section and apply it to additional use cases beyond decision tree ensembles.


Let us start by building some intuition about the structure of the $M$ matrices. A simple way to explore this structure is to plot the matrices themselves. \cref{fig:M_plot} shows examples of $M$ matrices generated by a single root-to-leaf path. These matrices exhibit a visually striking and highly structured pattern with numerous symmetries. For example, all non‑zero elements (the blue and red cells) form a Sierpinski triangle~\cite{sierpinski1915, decarli2025sierpinskitrianglegeneralizations}. This emerges because the \emph{MapPatternsToCube} algorithm does not add a cube in the $d[2p_c + 0][2p_b + 0]$ case (a structure analogous and symmetric to in~\cite[Exercise 1.2.4]{Topology_book}).

The Strassen matrix multiplication algorithm recursively partitions each matrix into four equally sized quadrants at every recursion step. 
Let $M$ be a $2^D \times 2^D$ matrix generated by the \woodelf algorithm, as discussed in~\cref{sec:woodelf_overview}. 
Consider a sub-matrix $M_{\text{sub}}$ encountered at recursion depth $i$ ($1 \le i \le D$). 
At this level of recursion, the sub-matrix $M_{\text{sub}}$ has dimensions $2^{D - i + 1} \times 2^{D - i + 1}$. 
All row indices within $M_{\text{sub}}$ share a common $(i-1)$-bit prefix $r \in \{0,1\}^{i-1}$, and all column indices share a common $(i-1)$-bit prefix $c \in \{0,1\}^{i-1}$. 

Let us further partition $M_{\text{sub}}$ into quarters:
\begin{enumerate}
    \item $M_1 = M_{\text{sub}}[\,c\boldsymbol{0}\dots][\,r\boldsymbol{0}\dots]$
    \item $M_2 = M_{\text{sub}}[\,c\boldsymbol{1}\dots][\,r\boldsymbol{0}\dots]$
    \item $M_3 = M_{\text{sub}}[\,c\boldsymbol{0}\dots][\,r\boldsymbol{1}\dots]$
    \item $M_4 = M_{\text{sub}}[\,c\boldsymbol{1}\dots][\,r\boldsymbol{1}\dots]$
\end{enumerate}

These matrices have useful mathematical properties:
\begin{itemize}
    \item $M_1 = 0$, since all its row and column indices have a $0$ at position $i$, and \emph{MapPatternsToCube} does not add a cube in the $d[2p_c + 0][2p_b + 0]$ case.
    
    \item $M_4 = M_2 + M_3$, due to the linearity of $v$.

    Let $M_2'$, $M_3'$, and $M_4'$ denote the matrices produced by \emph{MapPatternsToCube} before applying $v$:
    \[
        M_2 = v(M_2'), \qquad 
        M_3 = v(M_3'), \qquad 
        M_4 = v(M_4').
    \]

    For any two integers $0 \le g,h \le 2^{D - i - 1}$, all literals in the cubes at 
    $M_2'[g][h]$, $M_3'[g][h]$, and $M_4'[g][h]$ are identical except for the literal of $x_i$.  
    From lines~\ref{line:m2_creation}--\ref{line:m4_creation} of \cref{alg:fill_wdnf_table}, we know that 
    $x_i \in M_2'[g][h]$ (line~\ref{line:m2_creation}), 
    $\neg x_i \in M_3'[g][h]$ (line~\ref{line:m3_creation}), 
    and $x_i, \neg x_i \notin M_4'[g][h]$ (line~\ref{line:m4_creation}).  
    Thus:
    \[
        M_4'[g][h] = M_2'[g][h] + M_3'[g][h].
    \]

    Since $v$ is linear,
    \[
        v(M_4'[g][h]) = v(M_2'[g][h]) + v(M_3'[g][h]),
    \]
    and therefore $M_4 = M_2 + M_3$, as required.
\end{itemize}

These properties apply to $M$ matrices constructed using the Shapley value, Banzhaf value~\cite{banzhaf1965}, Shapley interaction values, and any other function that satisfies linearity.

We use these properties to propose a Strassen‑like matrix–vector multiplication that requires only two inner multiplications instead of four. This reduces the computational complexity to $O(2^D D)$. The approach is straightforward: see \cref{fig:strassen_formulation} for details.


A key observation is that the two inner multiplications depend only on $M_2$ and $M_3$, and do not use $M_1$ or $M_4$. 
Following this observation throughout the recursion, we see that the multiplication depends solely on the secondary diagonal of $M$ (the diagonal running from the top-right to the bottom-left). 
To save both space and computation time, we therefore compute and store only this secondary diagonal.

When constructing the secondary diagonal, we ignore all cases in which the row and column indices share the same bit value in any position. The \emph{MapPatternsToCube} algorithm (\cref{alg:fill_wdnf_table}) handles these cases at line~\ref{line:m4_creation}. Omitting this line yields the \emph{CubesInDiagonal} algorithm (\cref{alg:fill_wdnf_table}'), which computes only the cubes on the secondary diagonal.

\begin{figure}[t]
\centering
\includegraphics[width=0.99\linewidth]{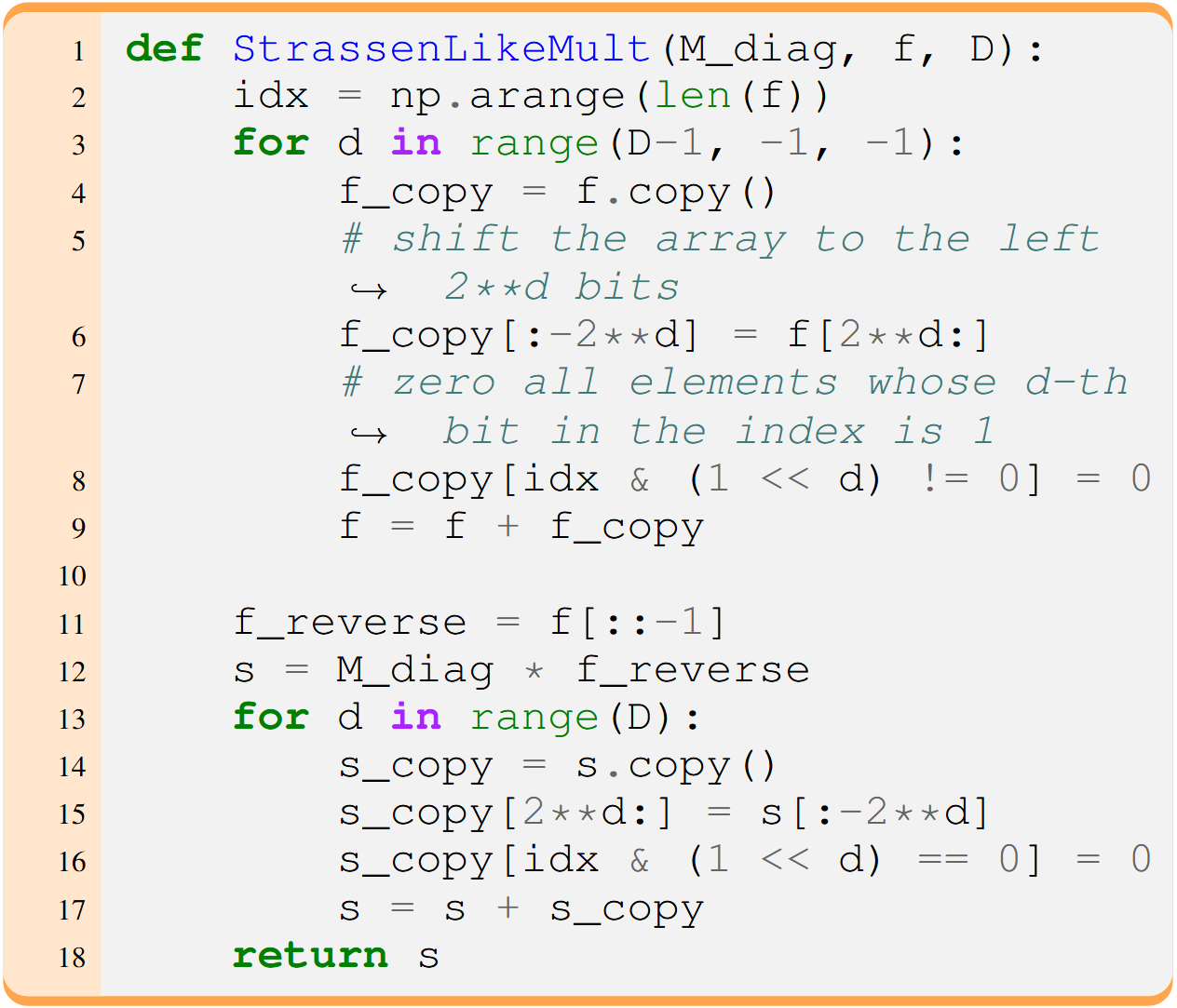}
\caption{
Python implementation of \emph{StrassenLikeMult}. 
Here, $M_{\text{diag}}$ is a NumPy array of length $2^D$ containing the secondary diagonal of $M$, 
$f$ is a NumPy array of length $2^D$, and $D$ is the tree depth. 
The function computes $M \cdot f$, see \cref{sec:vec_mult} for intuition and further details.
}
\label{fig:strassen_like_mult}
\end{figure}

Another improvement of our work is a fully vectorized implementation of the proposed multiplication scheme. The complete Python code is shown in \cref{fig:strassen_like_mult}, which also serves as a formal definition of the \emph{StrassenLikeMult} algorithm. Our implementation avoids recursion entirely and requires only $O(D)$ vectorized operations. Intuition and further details are provided in \cref{sec:vec_mult}.

    

\begin{figure*}[t]
\centering
\includegraphics[width=0.8\textwidth]{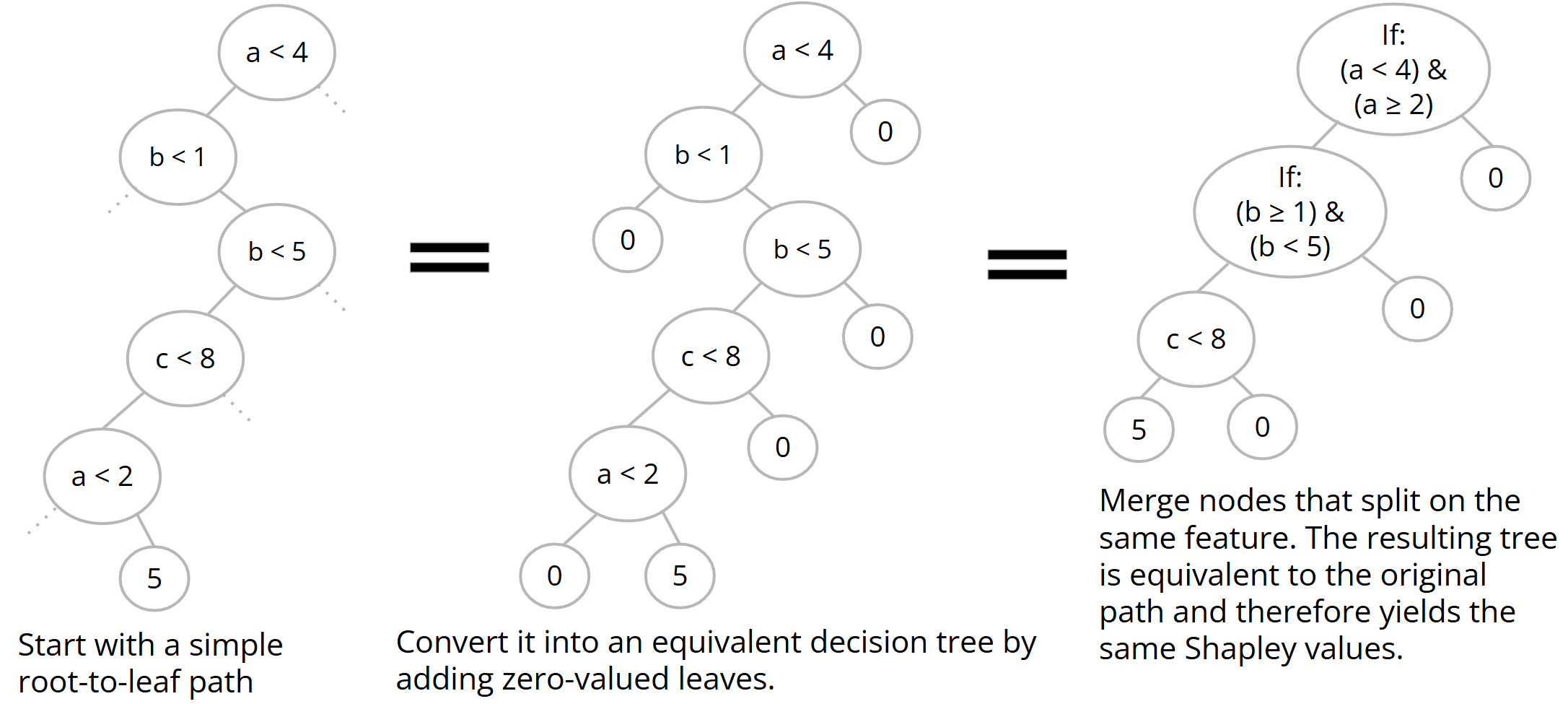} 
\caption{
Mathematical justification for merging all same‑feature splits into a single constraint.
}
\label{fig:merging_splits}
\end{figure*}

\section{Merging Same-Feature Splits}
\label{sec:merging_nodes}

In this section, we introduce a modified definition of decision patterns in which all nodes that split on the same feature are merged into a single constraint. This idea is inspired by a similar node‑merging technique used in \gputreeshap (\cite[Sect.~3.2]{GPUTreeShap}), but here it is incorporated directly into the formal definition of decision patterns. 

Using this formulation, we derive an algorithm for computing decision patterns that significantly reduces the space complexity compared to \woodelf. For Background SHAP, for example, it reduces the space complexity from $O(L3^DD+n(f+L)+mL)$ to $O(3^DD+n(f+D)+mD)$, see notations in \cref{fig:notations} (storing only the secondary diagonal of the $M$ matrix, as discussed in \cref{sec:strassen_mult}, further reduces the $3^D$ component to $2^D$).  

An additional advantage of this representation is that it enables a simple and compact caching mechanism that accelerates preprocessing. The mathematical justification for this approach is illustrated in \cref{fig:merging_splits}.

\paragraph{The Unique Features Decision Pattern (UFDP) (\cref{def:unique_features_decision_pattern})} 
for a leaf $l$ represents the decisions along the root-to-leaf path by assigning one bit to each feature that appears on the path. 
A bit is set to $1$ if all splits on that feature follow the path to $l$, and to $0$ if at least one such split diverges from it.
\begin{definition}[Unique Features Decision Pattern (UFDP)] 
\label{def:unique_features_decision_pattern}
Let $p$ be a decision pattern on a root-to-leaf path $(n_1 \equiv r_T, \dots, n_{D-1}, l)$. 
Let $k$ be the number of unique features appearing on this path, and let $f_1, \dots, f_k$ denote these features in the order of their first appearance
(e.g., $f_1$ is the feature used at the root). 
For each $f_i$, let $n_{i_1}, \dots, n_{i_{h_i}}$ be the nodes on the path that split on feature $f_i$.

The UFDP of the path is a binary sequence $u \in \{0,1\}^k$, such that 
its $i$-th bit is obtained from the decision pattern $p$ by:
\[
    u[i] = \bigwedge_{\forall i_t \in \{\,i_1, \dots, i_{h_i}\,\}} p[i_t] .
\]
\end{definition}

To implement \woodelf using UFDPs instead of standard decision patterns, we require a modified computation of the $f$ vectors. 
This replaces the original $f$-computation procedure described in~\cref{sec:woodelf_overview}.
For Background SHAP, the $f$ vector is obtained simply by applying \textit{value\_counts} to the UFDP patterns. 
For Path-Dependent SHAP, the $f_{l_{pd}}$ vector of leaf $l$ is computed by mapping each possible UFDP $u_b$ to its corresponding value, as defined below using the notation of~\cref{def:unique_features_decision_pattern}.
\begin{equation}
\label{eq:ufdp_path_dependent_estimation}
\mathbf{f_{l_{pd}}}[u_b] = \prod_{i=1}^{k} 
\begin{cases}
\prod_{t=1}^{h_i} \frac{n_{i_t+1}.\text{cover}}{n_{i_t}.\text{cover}} 
    & \text{if } u_b[i] = 1, \\[0.5em]
1 - \prod_{t=1}^{h_i} \frac{n_{i_t+1}.\text{cover}}{n_{i_t}.\text{cover}} 
    & \text{if } u_b[i] = 0 .
\end{cases}
\end{equation}

\paragraph{The \emph{DecisionPatternsGenerator} algorithm} efficiently computes UFDPs and replaces the \emph{CalcDecisionPatterns} algorithm used in \woodelf. 
The idea is straightforward: traverse the path from the root to the leaf, and at each node examine its splitting feature. 
Two cases may arise:
\begin{enumerate}
    \item If this feature has not been encountered before, append a new bit to the pattern with the current split result.
    \item If the feature has already been encountered, locate its existing bit in the pattern and apply a logical AND between that bit and the current split result.
\end{enumerate}

An additional advantage of the new approach is its improved space complexity. 
The algorithm is implemented as a generator that produces patterns leaf by leaf. 
This allows the system to limit storage to $O(D)$ nodes by immediately discarding the patterns of nodes whose children have all been processed. 
This represents a substantial improvement over \emph{CalcDecisionPatterns}, which stores the patterns of all nodes in the tree. 
Let $K$ denote the number of nodes in the tree (for a balanced tree, $K=L$). With $n$ decision patterns per node, the space complexity is reduced from $O(nK)$ to $O(nD)$. 
The pseudocode of \emph{DecisionPatternsGenerator} is provided in~\cref{appendix:DecisionPatternsGenerator}.

\paragraph{A Cache Mechanism:}

When using UFDPs instead of standard decision patterns, the input to \emph{CubesInDiagonal} becomes a list of unique features. Moreover, \emph{CubesInDiagonal} is independent of the specific identities of these features: for any root-to-leaf path with $k$ unique features, it produces the same dictionary $d_l$ and constructs the same set of $k$ matrices. Consequently, for a tree of maximum depth $D$, there are only $D$ distinct cases to handle -- one for each possible number of unique features. The \emph{ComputeMs} algorithm (\cref{alg:m_cache}) leverages this fact by precomputing all $D$ cases in advance and storing them in a compact cache.

\begin{algorithm}[t]
\caption{Compute and cache all possible $M$s}
\label{alg:m_cache}
\begin{algorithmic}[1]
\Function{ComputeMs}{$D$}
    \State $Ms = \{\}$
    \For{$k$ in $1,\dots,D$}
        \State $d = \textit{CubesInDiagonal}([1,\dots,k])$
        \For{$index$ in $d$}
            \State $cube = d[index]$
            \For{$feature, value$ in $v(cube)$}
                \State $Ms[k][feature][index] = value$
            \EndFor
        \EndFor
    \EndFor
    \State \Return $Ms$
\EndFunction
\end{algorithmic}
\end{algorithm}

\section{The \hdwoodelf Algorithm}

The original \woodelf algorithm executes the four logical steps described in~\cref{sec:woodelf_overview} independently for each tree, storing in memory for every leaf its corresponding $f$ vectors, $M$ matrices, and $s$ vectors. At peak memory usage, the algorithm maintains $D$ matrices of size $3^D$ and $2D$ vectors of size $2^D$ per leaf, resulting in a space complexity of $O(L\,3^D D)$.

In contrast, \hdwoodelf processes each leaf independently. At any point, it keeps in memory only the $D$ matrices and the $2D$ vectors associated with the leaf currently being processed. Combined with the observation that only the secondary diagonal of $M$ needs to be computed and stored, this reduces the space complexity from $O(L 3^D D)$ to $O(2^D D)$.

The pseudocode for \hdwoodelf is given in~\cref{alg:hd_woodelf_code}. The algorithm takes as input a decision tree $T$, a consumer dataset $C$, a background dataset $B$, and a linear function $v$ such as the Shapley value function. It computes and returns the $v$-values (e.g., Shapley values) for all consumers.

\begin{algorithm}[t]
\caption{\hdwoodelf algorithm}
\label{alg:hd_woodelf_code}
\begin{algorithmic}[1]
\Function{WoodelfForHighDepth}{$T, C, B, v$}
    \State $values = \{\}$
    \State $cpg = DecisionPatternsGenerator(T, C)$
    \State $bpg = DecisionPatternsGenerator(T, B)$
    \State $M_{cache} = \textit{ComputeMs}(T.depth)$
    \For{$l \in L_T$} \Comment{For each leaf $l$}
    
        \If{$|B| > 0$} \label{line:f_begin}
        \Comment{Step 1, Background $f$}
        \State $p_b = bpg.next()$
        \State $f = p_b.\textit{value\_counts}(\textit{normalize}=True)$ \label{line:value_counts}
        \Else 
        \Comment{Step 1, Path Dependent $f$}
        \State Compute $f$ using $l$ and~\cref{eq:ufdp_path_dependent_estimation}
        \EndIf \label{line:f_end}

    
        \Statex \Comment{Step 2, fetch $M$ from cache}
        \State $path = root\_to\_leaf\_path(T, l)$
        \State $pf = unique((n.\textit{feature})_{\forall n \in path})$
        \State $M_{diag} = M_{cache}[\:|pf|\:]$
        \State $s = \{\}$ \label{line:s_start}  \Comment{Step 3, compute $s$} 
        \For{$i, feature$ in $enumerate(pf)$}
            \State $a = \textit{StrassenLikeMult}( M_{diag}[i], f)$
            \State $s[feature] = w_l \cdot a$ \label{line:matrix_mult}
        \EndFor \label{line:s_end}
        \State $p_c = cpg.next()$ \label{line:final_computation_start} \Comment{Step 4, the final values}
        \For{$feature$ in $s$}
            \State $values[feature] \mathrel{+}= s[feature][\:p_c \:]$ \label{line:np_indexing}
        \EndFor
    \EndFor 
    \State \Return $values$
\EndFunction
\end{algorithmic}
\end{algorithm}

\begin{table*}[t]
\centering
\renewcommand{\arraystretch}{1.2}
\begin{tabularx}{\textwidth}{l|l|c|c|c|c|c|c}
\hline
\textbf{Task} & \textbf{Algorithm} & \textbf{$d{=}6$} & \textbf{$d{=}9$} & \textbf{$d{=}12$} & \textbf{$d{=}15$} & \textbf{$d{=}18$} & \textbf{$d{=}21$} \\
\hline

\multirow{5}{*}{Background SHAP}
 & \shap                   & 33 days & 61 days & 97 days & 134 days & 173 days & 209 days \\
 & \textsc{pltreeshap}     & 186 sec & 39 min & 3.8 hrs & 12.8 hrs* & X & X \\
 & \woodelf                & 12 sec & 84 sec & 58 min & 36 hrs* & X & X \\
 & \hdwoodelf              & 14 sec & 47 sec & 104 sec & 285 sec & 24 min & 3 hrs  \\
 & \hdwoodelf GPU          & 18 sec & 44 sec & 99 sec & 171 sec & 4.6 min & 12 min \\
\hline

\multirow{5}{*}{\makecell{Background SHAP IV \\ on \num{10000} samples}}
 & \textsc{pltreeshap}     & 63 sec & 383 sec & 33 min & X & X & X \\
 & \woodelf                & 9 sec & 108 sec & 2.2 hrs & X & X & X \\
 & \hdwoodelf              & 9 sec & 32 sec & 111 sec & 17 min & 2.6 hrs & X \\
 & \hdwoodelf GPU          & 15 sec & 50 sec & 118 sec & 3.8 min & 12 min & X \\
\hline

\end{tabularx}
\caption{
Runtime of Background SHAP and Background SHAP IV as a function of tree depth for different algorithms, evaluated on the IEEE-CIS dataset. 
Columns $d{=}6$, $d{=}9$, etc.\ correspond to trees of depths 6, 9, and so on. 
``X'' indicates runs that failed due to excessive memory usage (\textgreater 50GB RAM). 
Background SHAP IV is not implemented in \shap and thus omitted. 
Runtime estimation details: entries marked with * were measured on a single tree and extrapolated. 
Running times for \shap were measured using a background dataset of size 10 and extrapolated.
}
\label{tab:depth_running_time}
\end{table*}

\section{Experimental Results}
\label{sec:experimental_results}

We implemented the proposed \hdwoodelf approach and integrated it into the \woodelf Python package. 
All notebooks used in our experiments are available in the \woodelf Experiments repository.

We compared the performance of \hdwoodelf, the original \woodelf algorithm, \texttt{PLTreeSHAP}, and the \shap package. 
To evaluate \woodelf at scale, we used a large and well-known tabular dataset: the IEEE-CIS\footnote{The IEEE-CIS dataset: \url{https://www.kaggle.com/c/ieee-fraud-detection}} fraud detection dataset from the Kaggle competition. 
This dataset is widely recognized in the literature, with related studies including~\cite{frauddatasetwork1, frauddatasetwork2, frauddatasetwork3, frauddatasetwork5, frauddatasetwork6}. 

In IEEE-CIS, $|test| = \num{118108}$, $|train| = \num{472432}$, and $|features| = \num{397}$ after one-hot encoding of categorical features (where $F$ denotes the total number of features after preprocessing). In our experiments, we use the training set as the background dataset and the test set as the consumer dataset.

In both cases, we trained a \lightgbm classifier with 100 trees of varying depths. 
To allow deep paths while maintaining a realistic overall tree size, we increased the maximum number of leaves to 2024 (from the default of 31) and set \texttt{min\_data\_in\_leaf = 500} to limit overfitting. 
All algorithms were executed sequentially.

Our experiments were conducted in Google Colab’s CPU environment with the high-memory option enabled, providing 50GB of RAM instead of the standard 12GB. 
GPU experiments used the Colab T4 runtime with the high-memory configuration enabled. 


This section focuses on Background SHAP, for which \hdwoodelf improves the state-of-the-art complexity. Results for Path-Dependent SHAP, along with experiments on three additional datasets, are provided in our HTML report\footnote{The full experimental results are available in an interactive report generated by our experimental comparison framework: \url{https://ron-wettenstein.github.io/TreeBranchMarks/benchmarks/reports/WoodelfhdMainExperiment.html}}. 
For Path-Dependent SHAP, \hdwoodelf outperforms both the \shap package and the \woodelf algorithm at depths up to 18, while at depths 21 and above, the \shap package achieves better performance. For Background SHAP, \hdwoodelf outperforms all competing methods across all datasets.

\cref{tab:depth_running_time} summarizes the runtime of the Background SHAP and Background SHAP interaction values tasks on the IEEE-CIS dataset across different tree depths. For clarity of presentation and to keep the experimental runtime manageable, we report results for depths in increments of 3 (i.e., 6, 9, 12, 15, 18, and 21).

As shown in the table, computing Background SHAP on a large background dataset using the \shap package is impractical. 
Even at small tree depths, the runtime reaches several days since the \shap algorithm time complexity depends on $O(mn)$, 
where $n$ is the number of explained instances and $m$ is the size of the background dataset. 
\pltreeshap and \woodelf reduce this to $O(m + n)$ and therefore achieve substantially better performance. Consistent with the results reported in~\cite{woodelf}, \woodelf is more efficient than \pltreeshap. At depth 6, \woodelf is slightly faster than \hdwoodelf due to overhead from the space-optimized \emph{DecisionPatternsGenerator}.

However, since both \pltreeshap and \woodelf have space and time complexities that depend on $O(3^D D)$, they struggle with deeper trees. At depths 12 and 15 the runtime increases sharply, and at higher depths the algorithms exceed the available 50GB of memory.

In contrast, \hdwoodelf scales to deeper trees while delivering significant speedups even at moderate depths. 
When computing Background SHAP on CPU for an ensemble of depth 12, our method achieves a $33\times$ speedup over the previous state-of-the-art; 
at depth 15, it reaches a $162\times$ speedup. 
For Background SHAP IV, the speedups are even more pronounced. 
GPU acceleration further enhances performance, particularly for deep trees. 

Overall, these results show that exact SHAP computation, long considered impractical for deep trees, is now efficient and scalable in practice.

\section{Conclusion}
\label{sec:conclusion}

In this paper, we introduced \hdwoodelf, an extension of the \woodelf SHAP algorithm for efficient handling of high‑depth trees. \hdwoodelf reduces the dependence on the tree depth $D$ from $O(3^D D)$ to $O(2^D D^2)$, significantly advancing the state-of-the-art for Background SHAP. On the IEEE-CIS dataset, \hdwoodelf achieves a $33\times$ speedup at depth 12 and a $162\times$ speedup at depth 15. Moreover, it enables the computation of Shapley values at depths 18 and 21, when running on standard environments, where previous methods fail due to excessive memory usage.

Although \hdwoodelf covers most practical settings, rare cases with even greater depths remain challenging.
The ultimate goal is either to discover an $O(m+n)$ Background SHAP algorithm with a complexity that is purely polynomial in $D$, enabling efficient Background SHAP computation at any depth, or to establish formal lower bounds that rule out such an algorithm, providing fundamental insight into the inherent computational limits of Background SHAP. Our results indicate that \hdwoodelf, through its Strassen-like multiplication scheme, provides a promising step toward this goal while already enabling practical computation across a wide range of real-world scenarios.

\bibliographystyle{plainnat}
\bibliography{sample}

\clearpage

\appendix

\begin{figure*}[t]
\centering
\renewcommand{\arraystretch}{1.4}
\begin{tabular}{cc|c|c|c|c|}

    && \multicolumn{2}{c|}{add $x_1$} & \multicolumn{2}{c|}{} \\ 
    && add $x_2$ &  & add $x_2$ &  \\\hline

\multirow{2}{*}{add $\neg x_1$} &add $\neg x_2$ & $\emptyset$ & $\emptyset$ & $\emptyset$ & $(\neg x_1 \wedge \neg x_2)$ \\\cline{2-6}
& & $\emptyset$ & $\emptyset$ & $(\neg x_1 \wedge x_2)$ & $(\neg x_1)$ \\\hline
\multirow{2}{*}{} &add $\neg x_2$ & $\emptyset$ & $(x_1 \wedge \neg x_2)$ & $\emptyset$ & $(\neg x_2)$ \\\cline{2-6}
& & $(x_1 \wedge x_2)$ & $(x_1)$ & $(x_2)$ & $()$ \\\hline
\end{tabular}
\caption{Example of the $M_{\text{cubes}}$ matrix for $x_1$ and $x_2$. }
\label{fig:mcubes_example}
\end{figure*}

\begin{figure*}[t]
\centering
\renewcommand{\arraystretch}{1.2}
\setlength{\tabcolsep}{6pt}

\resizebox{\textwidth}{!}{%
\begin{tabular}{ccc|c|c|c|c|c|c|c|c|}
 & & & \multicolumn{4}{c|}{add $x_1$} & \multicolumn{4}{c|}{} \\
 & & & \multicolumn{2}{c|}{add $x_2$} & \multicolumn{2}{c|}{} & \multicolumn{2}{c|}{add $x_2$} & \multicolumn{2}{c|}{} \\
 & & & add $x_3$ &  & add $x_3$ &  & add $x_3$ &  & add $x_3$ &  \\
\hline

\multirow{4}{*}{add $\neg x_1$}
 & \multirow{2}{*}{add $\neg x_2$} & add $\neg x_3$ 
 & $\emptyset$ & $\emptyset$ & $\emptyset$ & $\emptyset$
 & $\emptyset$ & $\emptyset$ & $\emptyset$ & $(\neg x_1\wedge\neg x_2\wedge\neg x_3)$ \\
\cline{3-11}
 & &
 & $\emptyset$ & $\emptyset$ & $\emptyset$ & $\emptyset$
 & $\emptyset$ & $\emptyset$ & $(\neg x_1\wedge\neg x_2\wedge x_3)$ & $(\neg x_1\wedge\neg x_2)$ \\
\cline{2-11}

 & \multirow{2}{*}{} & add $\neg x_3$ 
 & $\emptyset$ & $\emptyset$ & $\emptyset$ & $\emptyset$
 & $\emptyset$ & $(\neg x_1\wedge x_2\wedge\neg x_3)$ & $\emptyset$ & $(\neg x_1\wedge \neg x_3)$ \\
\cline{3-11}
 & &
 & $\emptyset$ & $\emptyset$ & $\emptyset$ & $\emptyset$
 & $(\neg x_1\wedge x_2\wedge x_3)$ & $(\neg x_1\wedge x_2)$ & $(\neg x_1\wedge x_3)$ & $(\neg x_1)$ \\
\hline

\multirow{4}{*}{}
 & \multirow{2}{*}{add $\neg x_2$} & add $\neg x_3$ 
 & $\emptyset$ & $\emptyset$ & $\emptyset$ & $(x_1\wedge\neg x_2\wedge\neg x_3)$
 & $\emptyset$ & $\emptyset$ & $\emptyset$ & $(\neg x_2\wedge\neg x_3)$ \\
\cline{3-11}
 & &
 & $\emptyset$ & $\emptyset$ & $(x_1\wedge\neg x_2\wedge x_3)$ & $(x_1\wedge\neg x_2)$
 & $\emptyset$ & $\emptyset$ & $(\neg x_2\wedge x_3)$ & $(\neg x_2)$ \\
\cline{2-11}

 & \multirow{2}{*}{} & add $\neg x_3$ 
 & $\emptyset$ & $(x_1\wedge x_2\wedge\neg x_3)$ & $\emptyset$ & $(x_1\wedge \neg x3)$
 & $\emptyset$ & $(x_2\wedge\neg x_3)$ & $\emptyset$ & $(\neg x_3)$ \\
\cline{3-11}
 & &
 & $(x_1\wedge x_2\wedge x_3)$ & $(x_1\wedge x_2)$ & $(x_1\wedge x_3)$ & $(x_1)$
 & $(x_2\wedge x_3)$ & $(x_2)$ & $(x_3)$ & $()$ \\
\hline
\end{tabular}
}

\caption{Example of the $M_{\text{cubes}}$ matrix for $x_1,x_2,x_3$.}
\label{fig:mcubes_example_d3}
\end{figure*}

\section{A Boolean Logic View of the $M$ Matrices}
\label[appendix]{sec:M_matix_def_over_PB_functions}

In this section, we define the \woodelf $M$ matrix using only pseudo-Boolean terminology. The definition relies solely on the notion of a pseudo-Boolean function (\cref{def:PB_function}) and the definition of a cube (\cref{def:cube}). It does not refer to Shapley values or decision trees. We believe this perspective is advantageous because it generalizes our results and may facilitate their application in other domains, such as Boolean optimization and logic.

\begin{definition}[PB Function] \label{def:PB_function}
A \emph{Pseudo Boolean function} is a function of the form
$F(x_1,\dots,x_h): \{0,1\}^h \to \mathbb{R}$.
\end{definition}

\begin{definition}[Positive and Negative Literal; Cube; WDNF] \label{def:cube}
A \emph{literal} is a Boolean variable $x_i$ (\emph{positive literal}) or its negation $\neg x_i$ (\emph{negated literal}).  
A \emph{cube} is a conjunction of literals.  
A \emph{Weighted Disjunctive Normal Form (WDNF)} formula represents a PB function as:
\[
F(x_1,\dots,x_h) = \sum_{k=1}^n w_k \cdot c_k(x_1,\dots,x_h),
\]
where each $c_k$ is a cube and $w_k \in \mathbb{R}$ is its weight.
\end{definition}

Next, we define the linearity property over WDNF:

\begin{definition}[Linearity]
\label{def:linearity}
A function $g$ over WDNF formulas satisfies \emph{linearity} if its value on a WDNF formula $F = \sum_{k=1}^n w_k \cdot c_k$ can be expressed as the weighted sum of its values on the individual cubes:
\[
g(F) = \sum_{k=1}^n w_k \cdot g(c_k)
\]
\end{definition}

Using this pseudo-Boolean terminology (\cref{def:PB_function,def:cube,def:linearity}), we first define the cube matrix $M_{cubes}$. This matrix will later be used to define the $M$ matrices used by \woodelf.
Given Boolean variables $x_1,\dots,x_D$ their $M_{cubes}$ is a $2^D \times 2^D$ matrix that can be constructed as follows:
\begin{enumerate}
    \item The boolean representation of the row index is treated as a list of positive literals indicators. If the $i$ most significant bit is 0 add the literal $x_i$ to the cube.
    \item The boolean representation of the column index is treated as a list of negated literals indicators. If the $i$ most significant bit is 0 add the literal $\neg x_i$ to the cube.
    \item Remove cubes that have both the positive and the negated literal of the same variable as they are unsatisfiable. 
\end{enumerate}

\cref{def:M_cubes} formalize these rule:

\begin{definition} \label{def:M_cubes}
Let $x_1,\dots,x_D$ be Boolean variables. For any $a,b \in \{0,1\}^D$, the cubes matrix $M_{cubes}$ is defined as follows (Notation: $a \lor b$ is a bitwise OR and $a_i$ is the $i$ most significant bit of $a$):
\begin{equation}    
\begin{split}
M_{cubes}[a][b] \;=\; ( \bigwedge_{i:\, a_i = 0} x_i \;\wedge\; \bigwedge_{i:\, b_i = 0} \neg x_i \\ \text{if} \;\; a \lor b = 2^D-1 \;\; \text{else} \;\; \emptyset \; )
\end{split}
\end{equation}

\end{definition}

By construction, $M_{cubes}$ includes all possible satisfiable cubes over the Boolean variables $x_1,\dots,x_D$. The $M_{cubes}$ for $D=2$ and $D=3$ are visualized in~\cref{fig:mcubes_example} and~\cref{fig:mcubes_example_d3}, respectively. 

We can now define the \woodelf $M$ matrix.

\begin{definition}[The \woodelf $M$ matrix]
\label{def:pb_M_matrix}
The \woodelf $M$ matrix is constructed by applying a linear function on a cube matrix.
For a cube matrix $M_{cubes}$, a function $v$ that satisfies linearity and any $a,b \in \{0,1\}^D$, a \woodelf $M$ matrix can be constructed as follows:
\[
M[a][b] \;=\; v(M_{{cubes}}[a][b])
\]
\end{definition}

The results and algorithms presented in the Strassen-like multiplication section (\cref{sec:strassen_mult}) apply to any $M$ matrix that satisfies \cref{def:pb_M_matrix}.








\section{Vectorized Strassen-Like Matrix-Vector Multiplication}
\label[appendix]{sec:vec_mult}

This section provides intuition and further details about the \emph{StrassenLikeMult} algorithm presented in \cref{fig:strassen_like_mult}.

In \cref{sec:strassen_mult}, we propose an efficient matrix--vector multiplication algorithm that exploits the structure of the \woodelf \(M\) matrices. By leveraging the identities \(M_1 = 0\) and \(M_4 = M_2 + M_3\), the computation is reduced from four sub-matrix multiplications to two at each level of the recursion (see \cref{fig:strassen_formulation}).

\begin{figure*}[t]
\centering
\includegraphics[width=0.99\textwidth]{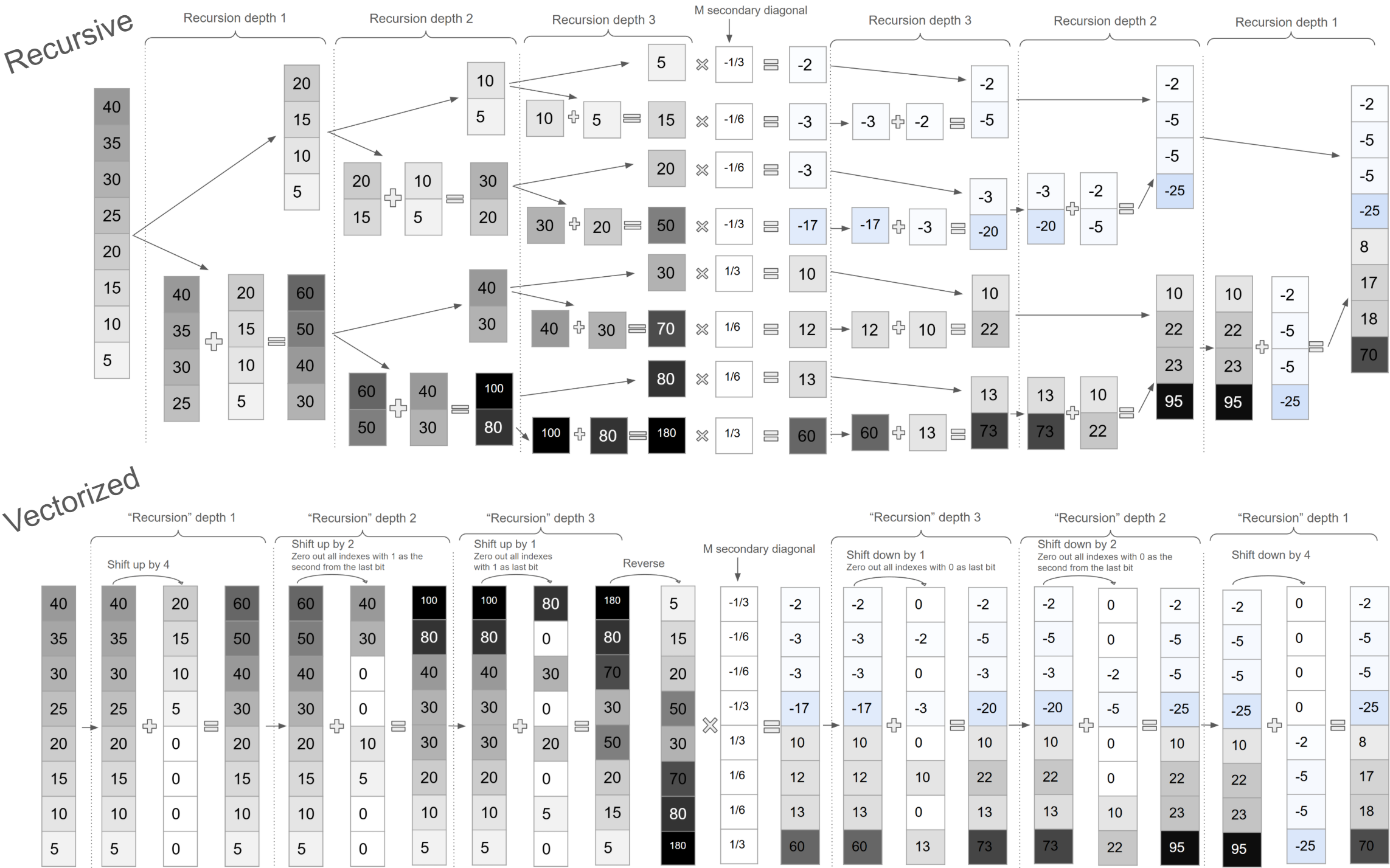} 
\caption{
Recursive and vectorized multiplication, as formalized by \cref{alg:recursive_mv} and \cref{fig:strassen_like_mult}, applied to a simple example. The figure illustrates how the vectorized approach performs all computations at a given recursion depth using just two steps: one operating on the input vector during the downward pass, and one operating on the resulting vector during the upward propagation.
}
\label{fig:vectorized_strassen_intuition}
\end{figure*}

\cref{alg:recursive_mv} presents a naive recursive implementation of this idea. While conceptually simple, it incurs significant overhead as the depth \(D\) increases, leading to \(2^D\) independent recursive calls. In contrast, our vectorized implementation processes all nodes at each recursion depth simultaneously using vectorized operations. As illustrated in \cref{fig:vectorized_strassen_intuition}, the vectorized computation can be viewed as two phases corresponding to the recursive downward and upward pass:
\begin{itemize}
    \item Downward Pass: We perform the $v_1 + v_2$ additions for every recursive depth in a single vectorized step using bit-masks and shifts.
    \item Upward Pass: After the core multiplication, propagate the results upward by aggregating the \(M_2 v_2\) and \(M_3 (v_1 + v_2)\) components using similar shift-based operations.
\end{itemize}

\cref{fig:vectorized_strassen_intuition} compares the recursive and vectorized executions side by side and highlights how each vectorized operation corresponds to a specific recursion depth. Although the theoretical complexity remains \(O(2^DD)\), the vectorized implementation is significantly faster in practice. It replaces an exponential number of recursive calls with only \(O(D)\) vectorized operations, which can be efficiently implemented using SIMD~\cite{numpy-nep38, numpy-simd-doc}.

\begin{algorithm}[t]
\caption{Recursive $M \cdot v$ Strassen-Like Multiplication}
\label{alg:recursive_mv}
\begin{algorithmic}[1]
\Function{MvRecursive}{$M, v$}
    
    \If{$M$ and $v$ are scalars}
        \State \Return $M \cdot v$
    \EndIf
    \Statex \Comment{Split $v$ to halves and $M$ to quarters.}
    \State $v_1, v_2 = v$
    \State $M_1, M_2, M_3, M_4 = M$
    \Statex \Comment{Assume $M_1=0$ and $M_2+M_3 = M_4$.}
    \Statex \Comment{Downward Pass:}
    \State $r_1 =$ \Call{MvRecursive}{$M_2, v2$}
    \State $r_2 =$ \Call{MvRecursive}{$M_3, v1+v2$}
    \Statex \Comment{Upward Pass:}
    \State $r = \textit{np.concat}([r_1, r_1+r_2])$
    \State \Return $r$
\EndFunction
\end{algorithmic}
\end{algorithm}

\section{DecisionPatternsGenerator Pseudo Code}
\label[appendix]{appendix:DecisionPatternsGenerator}

The \emph{DecisionPatternsGenerator} algorithm is discussed in~\cref{sec:merging_nodes}. It returns a dictionary $P$ that maps each leaf to the unique features decision patterns of its consumers, where $P[l_j][c_i]$ stores the pattern of consumer $c_i$ at leaf $l_j$. The same procedure can be applied to a background dataset, computing the pattern of each background row $b_i$ at leaf $l_j$. 

\cref{alg:decision_patterns_generator} presents the \emph{DecisionPatternsGenerator} algorithm pseudo code. We use the following notation: $n.l$ and $n.r$ denote the left and right child of the node $n$, respectively. For two integers $a,b$, the operation $a\&b$ denotes the bit-wise AND of their Boolean representation. 

The \emph{DecisionPatternsGenerator} algorithm replaces the \emph{CalcDecisionPatterns} algorithm used in \woodelf and significantly improves its space complexity. For reference, the pseudocode of \emph{CalcDecisionPatterns} is also provided in this section (see \cref{alg:calc_decision_patterns}). Note that \emph{CalcDecisionPatterns} computes decision patterns (\cref{def:decision_pattern}), whereas \emph{DecisionPatternsGenerator} computes unique-feature decision patterns (\cref{def:unique_features_decision_pattern}).

\begin{algorithm}[t]
\caption{Generate the UFDP for all leaves}
\label{alg:decision_patterns_generator}
\begin{algorithmic}[1]
\Function{DecisionPatternsGenerator}{$T, C$}
    \State $P \gets \{r_T: (0)_{\forall c \in C}\}$ 
    \State $F \gets \{r_T:[]\}$ \Comment{Map nodes to their path features}
    \For{$n$ in DFS(T)}: 
        \If{$n$ is a leaf}
            \State \Yield $P[n]$
        \Else
            \Statex \Comment{Update $F$}
            \If{$n.feature \notin F[n]$}
                \State $F[n.l] = F[n]+ [n.feature]$
                \State $F[n.r] = F[n]+ [n.feature]$
            \Else
                \State $F[n.l] = F[n]$
                \State $F[n.r] = F[n]$
            \EndIf

            \Statex \Comment{Update $P$ with $n.l$'s and $n.r$'s UFDPs}
            \State $s = n.\textit{split}(C)$
            \If{$n.feature \notin F[n]$}
                \State $P[n.l] = (P[n] << 1)
            + s$ 
                \State $P[n.r] = (P[n] << 1)
            + \lnot s$ 
            \Else
                \State $i = F[n].get\_index(n.feature)$
                \State $f_{bit} = 2^{|F[n]| - 1 - i}$
                \State $mask = 2^{|F[n]|-1} - f_{bit}$
                \State $P[n.l] = P[n] \: \& \: (mask + f_{bit} \cdot  s)$ 
                \State $P[n.r] = P[n]\:  \& \: (mask + f_{bit} \cdot  \lnot s)$ 
            \EndIf
        \EndIf 

        \Statex \Comment{When possible, clean up the RAM}
        \If{$n.parent.l \neq n$}
            \Statex \Comment{We can clean all the patterns of my left neighbor}
            \State $P.pop(n.parent.l)$
            \State $\textit{PopAllChildren}(P,n.parent.l)$
        \EndIf
    \EndFor
\EndFunction
\end{algorithmic}
\end{algorithm}

\begin{algorithm}[t]
\caption{{~\cite[Alg. 1]{woodelf}} Mapping Each Leaf to Its Decision Patterns}
\label{alg:calc_decision_patterns}
\begin{algorithmic}[1]
\Function{CalcDecisionPatterns}{$T, C$}
    \State $P_{\textit{leaves}} \gets \{\}$ \label{line:p_init}
    \State $P_{\textit{all}} \gets \{r_T: (0)_{\forall c \in C}\}$ \label{line:p_inner_nodes_init}
    \For{$n$ in BFS(T)}: \label{line:algo_bfs}
        \If{$n$ is a leaf}
            \State $P_{\textit{leaves}}[n] = P_{\textit{all}}[n]$
        \Else
            \State $P_{\textit{all}}[n.l] = (P_{\textit{all}}[n] << 1)
        + n.\textit{split}(C)$ 
            \State $P_{\textit{all}}[n.r] = (P_{\textit{all}}[n] << 1)
        + \lnot n.\textit{split}(C)$ 
           \label{line:inner_node_case_end}
        \EndIf \label{line:leaf_case_end}
    \EndFor
    \State \Return $P_{\textit{leaves}}$ \label{line:return_p}
\EndFunction
\end{algorithmic}
\end{algorithm}

\end{document}